\documentclass[a4paper]{article}

\usepackage{INTERSPEECH2021}
\usepackage{multirow}
\usepackage{xcolor}
\usepackage[hyphens]{url}

\title{Influence of ASR and Language Model on Alzheimer's Disease Detection} 
\name{Joan Codina-Filb\`a$^{1}$, Guillermo C\'ambara$^{1,2}$, Jordi Luque$^{2}$, Mireia Farr\'us$^{3}$}
\address{$^1$TALN Research Group, Universitat Pompeu Fabra, Barcelona, Spain\\
$^2$Telef\'onica I+D, Research, Barcelona, Spain\\
$^3$Language and Computation Centre, Universitat de Barcelona, Spain}
\email{joan.codina@upf.edu, guillermo.cambara@upf.edu, jordi.luque@telefonica.com, mfarrus@ub.edu}

\begin{document}

\maketitle
\begin{abstract}
  Alzheimer's Disease is the most common form of dementia. Automatic detection from speech could help to identify symptoms at early stages, so that preventive actions can be carried out. This research is a contribution to the ADReSS$_o$ Challenge, we analyze the usage of a SotA ASR system to transcribe participant's spoken descriptions from a picture. We analyse the loss of performance regarding the use of human transcriptions (measured using transcriptions from the 2020 ADReSS Challenge). Furthermore, we study the influence of a language model ---which tends to correct non-standard sequences of words--- with the lack of language model to decode the hypothesis from the ASR. This aims at studying the language bias and get more meaningful transcriptions based only on the acoustic information from patients. The proposed system combines acoustic ---based on prosody and voice quality--- and lexical features based on the first occurrence of the most common words. The reported results show the effect of using automatic transcripts with or without language model. The best fully automatic system achieves up to 76.06 $\%$ of accuracy (without language model), significantly higher, 3 $\%$ above, than a system employing word transcriptions decoded using general purpose language models.
  
\end{abstract}
\noindent\textbf{Index Terms}: Alzheimer's disease, prosody, voice quality features, lexical features, machine learning

\section{Introduction}

Dementia is a progressive syndrome characterised by a deterioration in cognitive function encompassing a severity spectrum, in which Alzheimer's Disease (AD) accounts about 70 \% of the cases. Its key symptoms, which can be manifested already in the early stages, are reflected by a loss of memory, a significant decline in thinking and reasoning skills, and language deterioration.

Language impairments related to AD are typically related to a progressive disruption in language integrity \cite{bayles1992relation}, with evidences of decline in performance at early stages \cite{mueller2018connected}. Language related features have been shown to capture the progression of language impairments; therefore, they have been proposed by several authors as markers of disease progression \cite{ahmed2013connected,beltrami2018speech}. Some studies have also reported the ability of prosodic and other linguistic features to classify different cognitive stages from speech, showing promising approaches for the identification of preclinical stages of dementia \cite{konig2018use,wang2019towards}. Moreover, it has been shown that AD subjects produce fewer specific words, while syntactic complexity, lexical complexity or conventional neuropsychological tests are not altered, suggesting that subtle spontaneous speech deficits might occur already in the very early stages of cognitive decline \cite{verfaillie2019high, CompanyW18}.

The current paper is part of the ADReSS$_o$ Challenge at Interspeech 2021 \cite{bib:LuzEtAl21ADReSSo}, which defines a shared task of AD detection based on a part of the Boston Diagnostic Aphasia Examination where users are asked to describe the Cookie Theft picture \cite{goodglass2001bdae}. With respect to the previous edition of the challenge (ADReSS 2020), ADReSS$_o$ (ADReSS, speech only) 2021 does not provide neither the manual transcriptions nor the part-of-speech (POS) and syntactic trees analyses, but only the speech files. However, the use of an automatic speech recognition (ASR) is allowed to get the text transcriptions. 

In our participation in the 2020 edition, our system achieved the second position in the detection of patients suffering AD \cite{farrus2020combining}, obtaining an accuracy of $87.5 \%$. The system combined prosodic, voice quality and lexical features obtained from manual transcriptions and POS tags. For the ADReSS$_o$ Challenge, we measure the effect, on performance, of replacing the manual transcriptions with an ASR, as well as the effect of removing any text processing (PoS taggers or syntactic analysis). In addition, we also measure the influence of using a language model in the ASR to see whether \emph{correcting} the output ASR transcriptions by means of a language model produces a loss of faithfulness with respect to the original speech of AD patients, so that important information might be lost on the way.

The structure of this paper unfolds as follows. Section \ref{soa} briefly overviews some systems to detect AD from speech, with special mention the ADReSS Challenge; section \ref{setup} describes the experimental setup of our work as well as the changes introduced for the ADReSS$_o$ Challenge, section \ref{results} compares the  obtained results, and finally, discussion and conclusions are drawn in sections \ref{discussion} and \ref{conclusions}, respectively.

\section{Alzheimer Disease detection from speech}\label{soa}

According to \cite{chien2019automatic}, systems relying on the speech signal to detect AD can be classified into those using acoustic-dependent features, and those using context-dependent features. The former systems rely mainly on extracting speech-based features regardless of the linguistic content ---i.e., the transcriptions---. These features can be related to spectral characteristics, voice quality, or prosody of speech, among others, and they can all be used in a language-independent way. In turn, prosody is conveyed through the intonation, stress, and rhythm elements, which are most robust over noise and channel distortions than, for instance, spectral features \cite{atal1972automatic,farrus2018voice}. Moreover, rhythm features such as speech rate can also be estimated without the need of the corresponding transcriptions \cite{de2009praat}. On the contrary, systems relying on context-dependent features require the word transcriptions to infer lexical, semantic, syntactic, and grammar information, among others. When manual transcriptions are not available, they can be obtained through ASR systems. Linguistically-based features can be very informative to detect AD \cite{balagopalan-etal-2020-impact,gosztolya2019identifying,zhou2016speech}; however, they rely on the performance of an ASR system, which are prone to introduce errors, or human transcriptions, which are very time-consuming, and they are language dependant. The more elaborate the language analysis is, more sensible to ASR errors will be, making a real system less robust.

Patterns in normal speech have been studied in some works \cite{farrus2016paragraph} and can serve as a basis for comparison in the analysis of acoustic information extracted from speech to detect AD status. For instance, \cite{luz2018method} makes use solely of acoustic-dependent features from spontaneous spoken dialogues, including prosody-related rhythm features such as speech rate and turn-taking patterns; others focus on spectral short-term features \cite{lopez2015automatic}, jointly with prosodic and voice quality features to combine it with an emotional speech analysis \cite{lopez2013selection}, or using standard acoustic parameters \cite{eyben2015geneva}. Voice quality features such as jitter and shimmer ---which account for cycle-to-cycle variations
of fundamental frequency and amplitude, respectively \cite{farrus2007jitter} --- and harmonicity, have also been used for AD detection \cite{meilan2014speech,mirzaei2017automatic,mirzaei2018two}. 

Apart from speech, semantic, lexical and grammar deficits in general have been reported as useful indicators \cite{salmon1988lexical,altmann2001speech}. In \cite{fritsch2019automatic}, the authors compute the perplexity of transliterations in the spoken description of the same picture to distinguish between AD and non-AD conditions. Similarly, \cite{wankerl2017n} performs an AD automatic diagnose based on n-gram models with subsequent evaluation of the perplexity, also from the transcriptions of spontaneous speech, and \cite{farzana2020exploring} uses the n-grams appearing at least five times. \cite{fraser2016linguistic} identifies several factors ---acoustic abnormality and impairments in semantics, syntax and information--- as linguistic markers in narrative speech to detect AD, and \cite{martinc2020tackling} uses readability features to assess AD. Some other recent works have also dealt with the use of pretrained word vectors \cite{cummins2020comparison,rohanian2020multi,edwards2020multiscale}.

\vspace{-1em}

\section{Experimental setup}\label{setup}

\subsection{Data}

This work has been carried out over two datasets, the one provided by the ADReSS challenge  in 2020, and the one provided by the ADReSS$_o$ Challenge . The ADReSS$_o$ dataset consists of 237 audio files, 166 for training and 71 for testing, balanced in both gender and AD/non-AD conditions, and with age ranging from 53 to 84 years old. In the ADReSS challenge there were 104 participants for training and 48 for testing.
All the participants, in both challenges, were asked to record a spoken description of the Cookie Theft picture from the Boston Diagnostic Aphasia Exam \cite{piccini1995natural,roth2011boston}. A more detailed description of the both datasets can be found in \cite{bib:LuzEtAl21ADReSSo} for ADReSS$_o$  and in   \cite{bib:LuzHaiderEtAl20ADReSS} for ADReSS. During the different experiments, the two datasets were not merged.

\subsection{Automatic Speech Recognition System} 

A pre-trained state-of-the-art ASR model from the wav2letter toolkit \cite{pratap2019wav2letter++} has been used to obtain the transcriptions from the participants' speech. The acoustic model of the ASR is based on the Transformer architecture proposed in \cite{synnaeve2019end}, incorporating convolutional operations to the Transformer layers \cite{vaswani2017attention}, the so-called Conformer blocks \cite{gulati2020conformer}. The model outputs directly any of the 26 English characters, plus the whitespace and the apostrophe. Regarding the lexicon, it consists of the top 200k words from the Common Crawl corpus. The Conformer acoustic model, which contains 300M parameters, plus the token set and the lexicon file are available at the wav2letter's RASR paper recipe\footnote{\url{https://github.com/facebookresearch/wav2letter/tree/master/recipes/rasr}} \cite{likhomanenko2020rethinking}. Such recipe has been chosen because it uses several datasets from different domains like read and conversational speech, for instance. This cross-domain representation aids the model to respond more robustly to the difficult spontaneous speech from the challenge. 

For the language model (LM), we have used the one provided with the MLS paper recipe\footnote{\url{https://github.com/facebookresearch/wav2letter/tree/master/recipes/mls}} \cite{pratap2020mls}, obtaining the best hypothesis via beam-search decoding. The model is a 5-gram LM trained with the KenLM toolkit, with around 133.26M sentences extracted from Project Gutenberg books. We have performed a grid-search of the beam-search hyper-parameters, by means of the ADReSS 2020 test set evaluation (which is not used in any of the evaluations performed afterwards), manually discarding some of the speakers utterances containing severe audio degradations, probably caused by speech enhancement. The outcome hyper-parameter choice from the grid-search can be seen in Table \ref{tab:lm_hyperparameters}, plus resulting Word Error Rate (WER) scores in ADReSS 2020 dataset at Table \ref{tab:asr_wers} .

\begin{table}[h]
  \caption{ASR language model hyper-parameters: language model weight (LM), word score (WS), silence score (SS), beam size (BS), beam threshold (BTH) and beam size token (BT)}
  \label{tab:lm_hyperparameters}
  \centering
  \begin{tabular}{cccccc}
    \toprule
    \textbf{LM} &  \textbf{WS} & \textbf{SS} & \textbf{BS} & \textbf{BTH} & \textbf{BT} \\
    \midrule
    $1.4$ & $1.5$  & $-0.2$ & $500$ & $1000$ & $30$ \\
    \bottomrule
  \end{tabular}
\vspace{-1.5em}
\end{table}

\begin{table}[h]
  \caption{ASR WER scores for ADReSS 2020 train and test sets, with and without LM}
  \label{tab:asr_wers}
  \centering
  \begin{tabular}{lcc}
    \toprule
    \textbf{Split Set} &  \textbf{WER w/ LM} & \textbf{WER w/o LM} \\
    \midrule
    Train & $44.98\%$  & $47.71\%$ \\
    Test & $42.89\%$  & $45.69\%$ \\
    \bottomrule
  \end{tabular}
\vspace{-1em} 
\end{table}

Although the use of an LM is recommended in terms of accuracy, it also might correct mispronunciations that could be interesting for dementia diagnosis. For such reason, we have chosen to generate, also, the transcriptions without LM to measure the effect of the LM itself. The generation without a LM is done applying Viterbi, decoding directly to the output scores from the acoustic model.

Since the output of Transformer-based models depends on the size of the input, we have split the audio according to the provided segmentation, to benefit from shorter segments. Then, we have concatenated back the transcriptions from consecutive participant segments, to obtain the text corresponding to a turn.

We have made our ASR recipe available for the community, so data processing and decoding scripts can be found at a GitHub repository\footnote{\url{https://github.com/gcambara/speechbook/tree/master/recipes/addresso}}.

\subsection{Feature extraction}

\subsubsection{Acoustic features}

We extracted a set of acoustic features using the Praat speech software \cite{boersma2018praat}, which we classified as prosodic (relying on intonation, stress, and rhythm), and voice quality features. To account for intonation, we extracted the following log-scaled F0-based features (measured in Hertz), using the cross-correlation method with an interval of 3.3 ms and a Gaussian window of length 80 ms: \emph{mean} and its \emph{std deviation}, \emph{max}, \emph{min}, \emph{range}, \emph{slope with octave jump removal}, and \emph{slope without octave jump removal}. It was also extracted: \emph{Mean intensity} (measured in dB), duration (in seconds) and rate features to account for rhythm: \emph{ratio of pauses}, \emph{average pause length}, \emph{speech rate}, \emph{articulation rate}, \emph{average syllable duration}, and \emph{effective duration of the speech}. Duration and rate features were based on the algorithm proposed by \cite{de2009praat} without using the provided transcriptions.

Voice quality features were also provided by Praat and included all jitter and shimmer measurements available (\emph{jitter\_loc}, \emph{jitter\_abs}, \emph{jitter\_rap}, \emph{jitter\_ppq5}, \emph{jitter\_ddp}, \emph{shimmer\_loc}, \emph{shimmer\_dB}, \emph{shimmer\_apq3}, \emph{shimmer\_apq5}, \emph{shimmer\_apq11}, and \emph{shimmer\_dda}), plus the following harmonicity-based features: \emph{harmonicity autocorrelation}, \emph{noise-to-harmonics ratio (NHR)}, and \emph{harmonics-to-noise ratio (HNR)}.

The audio files for each patient include a dialog with the interviewer (INV). In both challenges, the data provided includes the speaker identification. The patient segments were extracted from each file and the acoustic features computed for each segment. The different segments were aggregated using a weighted mean using the \emph{effective duration} as the weighing factor, while the \emph{effective duration} were added, not averaged. 

\subsubsection{Lexical features}

Lexical Features are based on transcriptions and some linguistic post processing. Treanscriptions can be manual (as in ADReSS) or automatic using an ASR (as suggested in ADReSS$_o$). In the ADReSS challenge, in addition to transcriptions, PoS tagging and syntactic trees were provided. For the 2021 challenge and ASR had to be used, and if required, perform the linguistic analysis over the extracted text. 

To participate the ADReSS challenge we extracted two types of features from the provided transcriptions: (1) the number of dialogue turns corresponding to the interviewer ---as an indicator of to what extent the subjects need further clarifications and stimuli---, min-max normalised, and (2) the first occurrence of the 50 most relevant words. Since all subjects performed the description of the same picture, we can consider this task as a summarizing task. Word position in text has been been used in many systems as a relevance measure for summarisation \cite{Hovy_97}. We expect that participants without any cognitive decline will catch the main ideas of the picture sooner, and their discourse will be structured from the most relevant to less relevant concepts. To produce the same feature vector for all the participants, and to keep the number of features of a reasonable size, we only considered the 50 top most frequent nouns, adjectives and verbs in the training set. For these words, for each participant, we computed their score based on their first occurrence in the dialog, using the following formula:
\begin{equation}
    S(j,w_i)=1-\frac{min(turn(j,w_i),maxTurns)}{maxTurns}
\end{equation}
where $S(j,w_i)$ is the score of word $i$ for user $j$; $turn(j,w_i)$ indicates the first turn where word $i$ is found ($\infty$ if not found), and $maxTurns$ the maximum number of turns in the dataset. 

In this challenge edition (ADReSS$_o$), we removed the stop words to get the list of most relevant words. Therefore, the system does not need any linguistic analysis (POS tagging or syntactic analysis). Turns are obtained considering the speaker provided by the challenge data. The ADReSS dataset was also processed in this way, to compare the the results obtained with or without transcriptions.

\subsection{Classification experiments}

As already mentioned, our system was performing very well on the ADReSS challenge, so our objective here was to measure the effect of using an ASR instead of human transcriptions in the classification accuracy. To do so, the next systems were trained and evaluated using 10 fold cross-validation (10-CV):
\begin{enumerate}
  \item ADReSS dataset
          \vspace{-0.3em}
        \begin{enumerate}
        \vspace{-0.3em}
            \item using Transcriptions (ADReSS challenge model)
                    \vspace{-0.3em}
            \item using Transcriptions using the same list of stopwords when using ASR.
                    \vspace{-0.3em}
            \item using ASR with Language Model   
            \vspace{-0.3em}
            \item using ASR without Language Model
                \vspace{-0.3em}
        \end{enumerate}
  \item ADReSS$_o$ dataset
          \vspace{-0.3em}
        \begin{enumerate}
                \vspace{-0.3em}
            \item using ASR with Language Model
                    \vspace{-0.3em}
            \item using ASR without Language Model
        \end{enumerate}
\end{enumerate}

The first model (1.a) was evaluated by the organisers of the ADReSS challenge with the provided test set, but this evaluation is not available anymore. Therefore, we were not able to compare the effect of using an ASR on the ADReSS test set. However, we can compare it with the results obtained using 10-CV on the train set. We also checked the effect of using stopwords to filter out some words instead of PoS labels. The models for the  ADReSS$_o$  dataset (2a,2b) were sent to be evaluated on the test set.

For the ADReSS challenge we used four different classifiers: Random Forest (RF), k-Nearest Neighbour, Support Vector Machines (SVM), and Multilayer Perceptron. Even that the best one was RF, the SVM was producing less false negatives for AD patients. We also found that SVM is more stable as the number of features is high (80) compared to the number of training samples in both datasets (104 and 166).    
To train the models we used the Support Vector Machines Classifier (SVC) implementation of Scikit-learn library in Python \footnote{https://scikit-learn.org/stable/}. (The code will be available for the final version of the paper). 


\section{Results}\label{results}

\begin{table}[tb]
  \caption{Accuracy of the different models depending on the data and transcriptions used. Test evaluation for ADReSS was not available for new models.} 
  \label{tab:classification1}
  \centering
\begin{tabular}{lllr}
  \toprule
 & Text  source & Train & Test \\
 \midrule
  \multirow{4}{*}{ADReSS  2020} &  Manual Transcr.  & 83.5\% & 81.2\% \\
  \multirow{4}{*} &  Manual Transcr. No PoS & 85.0\% & -- \\
  \multirow{4}{*} & ASR with LM & 77.8 \%& -- \\
  \multirow{4}{*} & ASR without LM & 72.3\% & -- \\
\midrule
 \multirow{2}{*}{ADReSS$_o$ 2021} & ASR with LM & 70.2\% & 73.2\% \\
 \multirow{2}{*}& ASR without LM & 73.3\% & 76.0\% \\
 \bottomrule
\end{tabular}
\vspace{-1.5em}
\end{table}

For this challenge, we participated in the binary classification task, that is, whether a patient is diagnosed with AD or not. 
Table \ref{tab:classification1} shows the accuracy of the predictions obtained by the different trained models on the different train sets, and for some of them, the results obtained on the corresponding challenge test. In all the models we use all the features. 
Since the datasets are different between both challenges, the results are not directly comparable. The results of the ADReSS challenge allow to measure the decrease in accuracy when the ASR replaces the human transcriptions. It can only be measured in the train set.
The models trained with the ADReSS$_o$ challenge allow us to better understand the effect of using a Language Model in the ASR pipeline, corroborated not only on training but also on the test set.

\vspace{-1em}

\section{Discussion}\label{discussion}

The analysis of the results in Table  \ref{tab:classification1} relative to the ADReSS challenge reveal that there is a drop in performance when the human transcriptions are replaced with the ones produced by an ASR system. The replacement of manual transcriptions by automatic ones also implies in our system the lack of POS tags. Without PoS tags, words are filtered using a stopwords list.

Surprisingly, performance increases (on the train set) on  $1.5\%$ when the words are filtered using a stopwords list instead of PoS tags, but maintaining the human transcriptions. The drop produced by using the ASR transcriptions, compared to using stopwords, ranges from $7,2 \%$ to a $12,7\% $. 

The results obtained with the ADReSS$_o$ dataset cannot be compared with human transcriptions, and we measure how the AD predictions are affected by the the use of a language model in the ASR. We observe that the model without LM outperforms by a $3 \%$ the one with LM, in both the train set (using 10-CV) and on the test set. This increase (which is inverted on the on the ADReSS dataset) made us wonder about the role of the LM when evaluating the mental decline and up to what extent is masking part of this decline by artificially biasing automatic transcripts produced by the ASR with an explicit language model.

We observed that the ASR (trained on a general purpose corpus) is giving a very high error rate (on the ADReSS dataset)  measured against the provided human transcriptions. We attribute part of this error rate to the sound quality of the audio files, another source of error can be the age of the population (elderly people) and finally their cognitive decline. Test set decoding yields a WER of $42.89\%$ when using a language model, while the same ASR on different corpora obtains WER scores smaller than $16 \%$ \cite{likhomanenko2020rethinking}.  This error rate increases to  $45.69 \%$ when not using a LM. We also observed that there is  a significant correlation between the Mini-Mental State Examination index (MMSE) and the WER, so that transcriptions from users with higher deterioration are more error prone. The mean WER of AD users is $57 \%$ while the one of control users is $29 \%$. Of course, when doing manual transcriptions the transcriber not only applies a language model to produce a meaningful text, it also uses context information (for example about the image being described) to guess the words corresponding to phonemes that does not fully understand.

ASR errors do not affect the list of most common words used as features. Both lists are very similar, and the most prominent differences are some pause fillers like \emph{'hum'} or \emph{'ah'}.

\vspace{-0.5em}

\section{Conclusions and future work}\label{conclusions}

In this paper we have studied the effect of using an ASR when performing an automatic detection system for AD using a set of acoustic and linguistic features. We used as a reference our system that was the second in performance on the ADReSS challenge. We discuss the effect of replacing manual transcriptions with ASR output, and the effect of using a language model as part of the ASR pipeline.

The system proposed is intentionally kept simple: it uses a small set of features 80, (30 acoustic, including prosodic, features and the relative location in the speech of the 50 most relevant words). We think that with a small dataset (166 participants) complex models may  produce overfitting. 

Acoustic-based systems have the advantage of being language-independent. Although we use ASR to extract words and we use turn-taking, this could also be done without the ASR, obtaining turn-taking by means of speaker diarisation, while the set of lexical words that are used as features, once the system is trained, could be detected by word spotting.

We think that the effect of using a LM on an ASR for detecting mental decline must be more deeply studied. As a future work, we want to visualize the corrections introduced by the LM and study if, and how, the LM is trying to mask part of this decline when trying to adjust the transcription to a coherent discourse.  

\vspace{-1em}

\section{Acknowledgements}

The fourth author has been funded by the Agencia Estatal de Investigación (AEI), Ministerio de Ciencia, Innovación y Universidades and the Fondo Social Europeo (FSE) under grant RYC-2015-17239 (AEI/FSE, UE).

\bibliographystyle{IEEEtran}

\bibliography{mybib}

\begin{thebibliography}{10}
\providecommand{\url}[1]{#1}
\csname url@samestyle\endcsname
\providecommand{\newblock}{\relax}
\providecommand{\bibinfo}[2]{#2}
\providecommand{\BIBentrySTDinterwordspacing}{\spaceskip=0pt\relax}
\providecommand{\BIBentryALTinterwordstretchfactor}{4}
\providecommand{\BIBentryALTinterwordspacing}{\spaceskip=\fontdimen2\font plus
\BIBentryALTinterwordstretchfactor\fontdimen3\font minus
  \fontdimen4\font\relax}
\providecommand{\BIBforeignlanguage}[2]{{%
\expandafter\ifx\csname l@#1\endcsname\relax
\typeout{** WARNING: IEEEtran.bst: No hyphenation pattern has been}%
\typeout{** loaded for the language `#1'. Using the pattern for}%
\typeout{** the default language instead.}%
\else
\language=\csname l@#1\endcsname
\fi
#2}}
\providecommand{\BIBdecl}{\relax}
\BIBdecl

\bibitem{bayles1992relation}
K.~A. Bayles, C.~K. Tomoeda, and M.~W. Trosset, ``Relation of linguistic
  communication abilities of {A}lzheimer's patients to stage of disease,''
  \emph{Brain and Language}, vol.~42, no.~4, pp. 454--472, 1992.

\bibitem{mueller2018connected}
K.~D. Mueller, B.~Hermann, J.~Mecollari, and L.~S. Turkstra, ``Connected speech
  and language in mild cognitive impairment and {A}lzheimer’s disease: A
  review of picture description tasks,'' \emph{Journal of Clinical and
  Experimental Neuropsychology}, vol.~40, no.~9, pp. 917--939, 2018.

\bibitem{ahmed2013connected}
S.~Ahmed, A.-M.~F. Haigh, C.~A. de~Jager, and P.~Garrard, ``Connected speech as
  a marker of disease progression in autopsy-proven {A}lzheimer’s disease,''
  \emph{Brain}, vol. 136, no.~12, pp. 3727--3737, 2013.

\bibitem{beltrami2018speech}
D.~Beltrami, G.~Gagliardi, R.~Rossini~Favretti, E.~Ghidoni, F.~Tamburini, and
  L.~Calz{\`a}, ``Speech analysis by natural language processing techniques: A
  possible tool for very early detection of cognitive decline?''
  \emph{Frontiers in Aging Neuroscience}, vol.~10, p. 369, 2018.

\bibitem{konig2018use}
A.~Konig, A.~Satt, A.~Sorin, R.~Hoory, A.~Derreumaux, R.~David, and P.~H.
  Robert, ``Use of speech analyses within a mobile application for the
  assessment of cognitive impairment in elderly people,'' \emph{Current
  Alzheimer Research}, vol.~15, no.~2, pp. 120--129, 2018.

\bibitem{wang2019towards}
T.~Wang, C.~Lian, J.~Pan, Q.~Yan, F.~Zhu, M.~L. Ng, L.~Wang, and N.~Yan,
  ``Towards the speech features of mild cognitive impairment: Universal
  evidence from structured and unstructured connected speech of {C}hinese,'' in
  \emph{Proc. Interspeech}, 2019.

\bibitem{verfaillie2019high}
S.~C. Verfaillie, J.~Witteman, R.~E. Slot, I.~J. Pruis, L.~E. Vermaat
  \emph{et~al.}, ``High amyloid burden is associated with fewer specific words
  during spontaneous speech in individuals with subjective cognitive decline,''
  \emph{Neuropsychologia}, vol. 131, pp. 184--192, 2019.

\bibitem{CompanyW18}
J.~{Soler-Company} and L.~Wanner, ``Automatic identification of texts written
  by authors with {A}lzheimer's disease,'' in \emph{Proceedings of the 40th
  Annual Meeting of the Cognitive Science Society, CogSci 2018, USA, July
  25-28}, 2018.

\bibitem{bib:LuzEtAl21ADReSSo}
S.~Luz, F.~Haider, S.~de~la Fuente, D.~Fromm, and B.~MacWhinney, ``Detecting
  cognitive decline using speech only: The adresso challenge,'' \emph{medRxiv},
  2021.

\bibitem{goodglass2001bdae}
H.~Goodglass, E.~Kaplan, and S.~Weintraub, \emph{BDAE: The Boston Diagnostic
  Aphasia Examination}.\hskip 1em plus 0.5em minus 0.4em\relax Lippincott
  Williams \& Wilkins Philadelphia, PA, 2001.

\bibitem{farrus2020combining}
M.~Farrús and J.~Codina-Filbà, ``Combining prosodic, voice quality and
  lexical features to automatically detect {A}lzheimer's disease,'' \emph{arXiv
  preprint arXiv:2011.09272}, 2020.

\bibitem{chien2019automatic}
Y.-W. Chien, S.-Y. Hong, W.-T. Cheah, L.-H. Yao, Y.-L. Chang \emph{et~al.},
  ``An automatic assessment system for {A}lzheimer’s disease based on speech
  using feature sequence generator and recurrent neural network,''
  \emph{Scientific Reports}, vol.~9, no.~1, pp. 1--10, 2019.

\bibitem{atal1972automatic}
B.~S. Atal, ``Automatic speaker recognition based on pitch contours,''
  \emph{The Journal of the Acoustical Society of America}, vol.~52, no.~6B, pp.
  1687--1697, 1972.

\bibitem{farrus2018voice}
M.~Farr{\'u}s, ``Voice disguise in automatic speaker recognition,'' \emph{ACM
  Computing Surveys (CSUR)}, vol.~51, no.~4, pp. 1--22, 2018.

\bibitem{de2009praat}
N.~H. De~Jong and T.~Wempe, ``Praat script to detect syllable nuclei and
  measure speech rate automatically,'' \emph{Behavior Research Methods},
  vol.~41, no.~2, pp. 385--390, 2009.

\bibitem{balagopalan-etal-2020-impact}
A.~Balagopalan, K.~Shkaruta, and J.~Novikova, ``Impact of {ASR} on
  {A}lzheimer{'}s disease detection: All errors are equal, but deletions are
  more equal than others,'' in \emph{Proc. of the Sixth Workshop on Noisy
  User-generated Text}, 2020.

\bibitem{gosztolya2019identifying}
G.~Gosztolya, V.~Vincze, L.~T{\'o}th, M.~P{\'a}k{\'a}ski, J.~K{\'a}lm{\'a}n,
  and I.~Hoffmann, ``Identifying mild cognitive impairment and mild
  alzheimer’s disease based on spontaneous speech using asr and linguistic
  features,'' \emph{Computer Speech \& Language}, vol.~53, pp. 181--197, 2019.

\bibitem{zhou2016speech}
L.~Zhou, K.~C. Fraser, and F.~Rudzicz, ``Speech recognition in {A}lzheimer's
  disease and in its assessment.'' in \emph{Interspeech}, 2016, pp. 1948--1952.

\bibitem{farrus2016paragraph}
M.~Farr{\'u}s, C.~Lai, and J.~D. Moore, ``Paragraph-based prosodic cues for
  speech synthesis applications,'' in \emph{Proc. Interspeech}, pp. 1143--7.

\bibitem{luz2018method}
S.~Luz, S.~de~la Fuente, and P.~Albert, ``A method for analysis of patient
  speech in dialogue for dementia detection,'' in \emph{Proceedings of LREC},
  2018.

\bibitem{lopez2015automatic}
K.~Lopez-de Ipi{\~n}a, J.~B. Alonso, J.~Sol{\'e}-Casals, N.~Barroso,
  P.~Henriquez, M.~Faundez-Zanuy, C.~M. Travieso, M.~Ecay-Torres,
  P.~Martinez-Lage, and H.~Eguiraun, ``On automatic diagnosis of
  {A}lzheimer’s disease based on spontaneous speech analysis and emotional
  temperature,'' \emph{Cognitive Computation}, vol.~7, no.~1, pp. 44--55, 2015.

\bibitem{lopez2013selection}
K.~L{\'o}pez-de Ipi{\~n}a, J.-B. Alonso, C.~M. Travieso, J.~Sol{\'e}-Casals,
  H.~Egiraun, M.~Faundez-Zanuy, A.~Ezeiza, N.~Barroso, M.~Ecay-Torres,
  P.~Martinez-Lage \emph{et~al.}, ``On the selection of non-invasive methods
  based on speech analysis oriented to automatic {A}lzheimer disease
  diagnosis,'' \emph{Sensors}, vol.~13, no.~5, pp. 6730--6745, 2013.

\bibitem{eyben2015geneva}
F.~Eyben, K.~R. Scherer, B.~W. Schuller, J.~Sundberg, E.~Andr{\'e}, C.~Busso,
  L.~Y. Devillers, J.~Epps, P.~Laukka, S.~S. Narayanan \emph{et~al.}, ``The
  geneva minimalistic acoustic parameter set (gemaps) for voice research and
  affective computing,'' \emph{IEEE {T}ransactions on {A}ffective {C}omputing},
  vol.~7, no.~2, pp. 190--202, 2015.

\bibitem{farrus2007jitter}
M.~Farr{\'u}s, J.~Hernando, and P.~Ejarque, ``Jitter and shimmer measurements
  for speaker recognition,'' in \emph{Proc. Interspeech}, 2007.

\bibitem{meilan2014speech}
J.~J.~G. Meil{\'a}n, F.~Mart{\'\i}nez-S{\'a}nchez, J.~Carro, D.~E. L{\'o}pez,
  L.~Millian-Morell, and J.~M. Arana, ``Speech in {A}lzheimer's disease: Can
  temporal and acoustic parameters discriminate dementia?'' \emph{Dementia and
  Geriatric Cognitive Disorders}, vol.~37, no. 5-6, pp. 327--334, 2014.

\bibitem{mirzaei2017automatic}
S.~Mirzaei, M.~El~Yacoubi, S.~Garcia-Salicetti, J.~Boudy, C.~K.~S. Muvingi,
  V.~Cristancho-Lacroix, H.~Kerherv{\'e}, and A.-S.~R. Monnet, ``Automatic
  speech analysis for early {A}lzheimer's disease diagnosis,'' in
  \emph{Proceedings of the 6e Journées d'Etudes sur la Télésanté}, 2017.

\bibitem{mirzaei2018two}
S.~Mirzaei, M.~El~Yacoubi, S.~Garcia-Salicetti, J.~Boudy, C.~Kahindo,
  V.~Cristancho-Lacroix, H.~Kerherv{\'e}, and A.-S. Rigaud, ``Two-stage feature
  selection of voice parameters for early {A}lzheimer's disease prediction,''
  \emph{IRBM}, vol.~39, no.~6, pp. 430--435, 2018.

\bibitem{salmon1988lexical}
D.~P. Salmon, A.~P. Shimamura, N.~Butters, and S.~Smith, ``Lexical and semantic
  priming deficits in patients with {A}lzheimer's disease,'' \emph{Journal of
  Clinical and Experimental Neuropsychology}, vol.~10, no.~4, pp. 477--494,
  1988.

\bibitem{altmann2001speech}
L.~J. Altmann, D.~Kempler, and E.~S. Andersen, ``Speech errors in {A}lzheimer's
  disease,'' \emph{Journal of Speech, Language, and Hearing Research}, 2001.

\bibitem{fritsch2019automatic}
J.~Fritsch, S.~Wankerl, and E.~N{\"o}th, ``Automatic diagnosis of
  {A}lzheimer’s disease using neural network language models,'' in
  \emph{ICASSP-2019}, 2019, pp. 5841--5845.

\bibitem{wankerl2017n}
S.~Wankerl, E.~N{\"o}th, and S.~Evert, ``An n-gram based approach to the
  automatic diagnosis of {A}lzheimer's disease from spoken language.'' in
  \emph{Proc. Interspeech}, 2017, pp. 3162--3166.

\bibitem{farzana2020exploring}
S.~Farzana and N.~Parde, ``Exploring mmse score prediction using verbal and
  non-verbal cues,'' \emph{Proc. Interspeech 2020}, pp. 2207--2211, 2020.

\bibitem{fraser2016linguistic}
K.~C. Fraser, J.~A. Meltzer, and F.~Rudzicz, ``Linguistic features identify
  {A}lzheimer’s disease in narrative speech,'' \emph{Journal of Alzheimer's
  Disease}, vol.~49, no.~2, pp. 407--422, 2016.

\bibitem{martinc2020tackling}
M.~Martinc and S.~Pollak, ``Tackling the adress challenge: a multimodal
  approach to the automated recognition of {A}lzheimer’s dementia,''
  \emph{Proc. Interspeech 2020}, pp. 2157--2161, 2020.

\bibitem{cummins2020comparison}
N.~Cummins, Y.~Pan, Z.~Ren, J.~Fritsch, V.~S. Nallanthighal, H.~Christensen,
  D.~Blackburn, B.~W. Schuller, M.~Magimai-Doss, H.~Strik \emph{et~al.}, ``A
  comparison of acoustic and linguistics methodologies for {A}lzheimer’s
  dementia recognition,'' in \emph{Proc. Interspeech 2020}, 2020, pp.
  2182--2186.

\bibitem{rohanian2020multi}
M.~Rohanian, J.~Hough, and M.~Purver, ``Multi-modal fusion with gating using
  audio, lexical and disfluency features for {A}lzheimer’s dementia
  recognition from spontaneous speech,'' in \emph{Proc. Interspeech}, 2020, pp.
  2187--2191.

\bibitem{edwards2020multiscale}
E.~Edwards, C.~Dognin, B.~Bollepalli, M.~Singh, and V.~Analytics, ``Multiscale
  system for {A}lzheimer’s dementia recognition through spontaneous speech,''
  \emph{Proc. Interspeech 2020}, pp. 2197--2201, 2020.

\bibitem{piccini1995natural}
C.~Piccini, L.~Bracco, M.~Falcini, G.~Pracucci, and L.~Amaducci, ``Natural
  history of {A}lzheimer's disease: prognostic value of plateaux,''
  \emph{Journal of the Neurological Sciences}, vol. 131, no.~2, pp. 177--182,
  1995.

\bibitem{roth2011boston}
C.~Roth, ``Boston diagnostic aphasia examination,'' \emph{Encyclopedia of
  Clinical Neuropsychology}, pp. 428--430, 2011.

\bibitem{bib:LuzHaiderEtAl20ADReSS}
S.~Luz, F.~Haider, S.~de~la Fuente, D.~Fromm, and B.~MacWhinney,
  ``{Alzheimer's} dementia recognition through spontaneous speech: The {ADReSS
  Challenge},'' in \emph{Proceedings of INTERSPEECH}, Shanghai, China, 2020.

\bibitem{pratap2019wav2letter++}
V.~Pratap, A.~Hannun, Q.~Xu, J.~Cai, J.~Kahn, G.~Synnaeve, V.~Liptchinsky, and
  R.~Collobert, ``Wav2letter++: A fast open-source speech recognition system,''
  in \emph{ICASSP-2019}.\hskip 1em plus 0.5em minus 0.4em\relax IEEE, 2019, pp.
  6460--6464.

\bibitem{synnaeve2019end}
G.~Synnaeve, Q.~Xu, J.~Kahn, T.~Likhomanenko, E.~Grave, V.~Pratap, A.~Sriram,
  V.~Liptchinsky, and R.~Collobert, ``End-to-end asr: from supervised to
  semi-supervised learning with modern architectures,'' \emph{arXiv preprint
  arXiv:1911.08460}, 2019.

\bibitem{vaswani2017attention}
A.~Vaswani, N.~Shazeer, N.~Parmar, J.~Uszkoreit, L.~Jones, A.~N. Gomez,
  L.~Kaiser, and I.~Polosukhin, ``Attention is all you need,'' \emph{arXiv
  preprint arXiv:1706.03762}, 2017.

\bibitem{gulati2020conformer}
A.~Gulati, J.~Qin, C.-C. Chiu, N.~Parmar, Y.~Zhang, J.~Yu, W.~Han, S.~Wang,
  Z.~Zhang, Y.~Wu \emph{et~al.}, ``Conformer: Convolution-augmented transformer
  for speech recognition,'' \emph{arXiv preprint arXiv:2005.08100}, 2020.

\bibitem{likhomanenko2020rethinking}
T.~Likhomanenko, Q.~Xu, V.~Pratap, P.~Tomasello, J.~Kahn, G.~Avidov,
  R.~Collobert, and G.~Synnaeve, ``Rethinking evaluation in asr: Are our models
  robust enough?'' \emph{arXiv preprint arXiv:2010.11745}, 2020.

\bibitem{pratap2020mls}
V.~Pratap, Q.~Xu, A.~Sriram, G.~Synnaeve, and R.~Collobert, ``Mls: A
  large-scale multilingual dataset for speech research,'' \emph{arXiv preprint
  arXiv:2012.03411}, 2020.

\bibitem{boersma2018praat}
P.~Boersma and D.~Weenink, ``Praat: Doing phonetics by computer [computer
  program]. version 6.0.37,'' \emph{Retrieved February}, 2018.

\bibitem{Hovy_97}
C.-Y. Lin and E.~Hovy, ``Identifying topics by position,'' in \emph{Proceedings
  of the Fifth Conference on Applied Natural Language Processing}.\hskip 1em
  plus 0.5em minus 0.4em\relax USA: ACL, 1997, p. 283–290.

\end{thebibliography}


\end{document}